\documentclass[conference]{IEEEtran}
\IEEEoverridecommandlockouts
\usepackage{cite}
\usepackage{amsmath,amssymb,amsfonts}
\usepackage{graphicx}
\usepackage{textcomp}
\usepackage{xcolor}
\usepackage{hyperref}
\usepackage{booktabs}
\usepackage{multirow}
\usepackage{algorithm,algpseudocode}
\usepackage{url}
\usepackage{balance}
\usepackage{caption}
\usepackage{subcaption}
\usepackage{eso-pic}

\usepackage{tikz} 
\usetikzlibrary{shapes.geometric, arrows, positioning}

\def\BibTeX{{\rm B\kern-.05em{\sc i\kern-.025em b}\kern-.08em
    T\kern-.1667em\lower.7ex\hbox{E}\kern-.125emX}}
\begin{document}

\title{A Generative Adversarial Augmentation Framework for Robust and Deployable DDoS Attack Detection}
\title {A GAN-Based Framework for Robust DDoS Attacks Detection}

\author{
    Makram Chehayeb$^1$, Walid Fahs$^{1,2}$, Amina Rizk$^2$, Rida Khatoun$^3$, Omran Berjawi$^3$ \\
    \small{$^1$Islamic University of Lebanon, Faculty of Engineering, Lebanon} \\
    \small{$^2$Lebanese University, Faculty of Sciences, Lebanon} \\
    \small{$^3$Polytechnic Institute of Paris, Télécom Paris, LTCI, France} \\[0.5em]

    \texttt{makram.chehayeb@iul.edu.lb, walid.fahs@iul.edu.lb, amina.rizk03@gmail.com}  \\
    \texttt{rida.khatoun@telecom-paris.fr, omran.berjawi@telecom-paris.fr}
}
\title{A GAN-Based Framework for Robust DDoS Attack Detection}
\maketitle

\AddToShipoutPictureBG*{%
  \AtPageLowerLeft{%
    \raisebox{0.6cm}{%
      \makebox[\paperwidth]{%
        \begin{minipage}{\textwidth}\centering\scriptsize
          \copyright~2026 IEEE. Personal use of this material is permitted.
          Permission from IEEE must be obtained for all other uses, in any
          current or future media, including reprinting/republishing this
          material for advertising or promotional purposes, creating new
          collective works, for resale or redistribution to servers or lists,
          or reuse of any copyrighted component of this work in other works.
        \end{minipage}%
      }%
    }%
  }%
}

\begin{abstract}
The availability and consistency of online services remain vulnerable due to Distributed Denial of Service (DDoS) attacks. These attacks are evolving by adopting more complex strategies to evade traditional network security systems. Despite the effectiveness of machine learning models in detecting DDoS traffic, targeted adversarial attacks can degrade their classification accuracy. This work proposes a robust detection framework that integrates generative adversarial modelling with advanced machine learning models. We trained Random Forests, Deep Neural Ensembles, and Transformer-based models using the CICDDoS2019 dataset to establish the framework’s baseline performance. To enhance the models’ defensive capacity, we generated synthetic adversarial flows that simulate potential evasion attempts and adversarial traffic using a Wasserstein Generative Adversarial Network with Gradient Penalty (WGAN-GP). Then, we combined the generated traffic with benign and malicious traffic to construct hybrid datasets to train the models to learn more generalizable decision boundaries. The experimental results indicate that the proposed methodology significantly enhances detection accuracy and resilience, especially against unseen adversarial traffic. We also tested the designed framework using real-world generated traffic, which demonstrates its capability in practical settings. The scalable and efficient solution against adversarial DDoS attacks, introduced in this work, paves the way towards more resilient and adaptive network defense systems that combine generative adversarial augmentation with recent advances in learning models.
\end{abstract}


\section{Introduction}
Nowadays, the increase in services being provided via cloud environments, the number of Internet of Things (IoT) devices connected to networks, and the availability of larger botnets allow attackers to conduct some Distributed Denial of Service (DDoS) campaigns that are larger and more complex, defeating or bypassing many legacy defenses. These attacks are critical threats that have critical economic consequences, and they highlight the vulnerabilities of online services that rely on consistent service uptime, such as remote medical tele-surveillance and smart home health monitoring. \cite{12,14}.

\begin{figure}[htbp]
    \centering
    \includegraphics[width=0.9\linewidth]{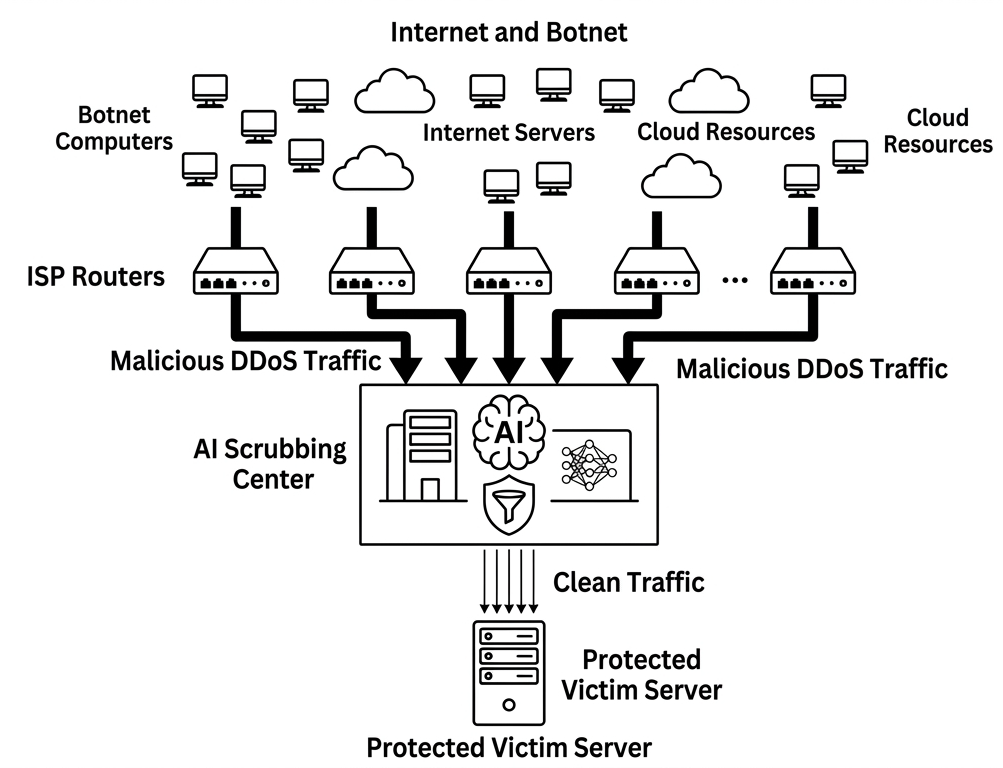}
    \caption{General architecture of AI-assisted DDoS mitigation using traffic scrubbing centers and intelligent filtering.}
    \label{fig:general_ddos}
\end{figure}

Due to the complexity and rising scale of modern cyberattacks, the DDoS protection domain has experienced a considerable expansion recently. These defensive solutions range from local physical appliances (on-premise) to massive cloud-based scrubbing services\cite{13}. This market is currently led by several key players, including Cloudflare, Akamai, NETSCOUT (Arbor), Radware, and AWS Shield. These providers employ a mixture of techniques, including rate limiting, behavioral analytics, and Artificial Intelligence (AI), to create a multi-level defense. Despite these defensive mechanisms, a significant gap persists: adversarial attacks engineered to bypass AI-based detection engines \cite{6}. Figure~\ref{fig:general_ddos} illustrates a typical AI-assisted DDoS mitigation workflow in which malicious traffic generated by distributed botnets is filtered through intelligent scrubbing infrastructure before reaching the protected server.

To overcome these challenges, machine learning-based Intrusion Detection Systems (IDS) have been widely adopted due to their capacity to map complex traffic patterns and detect anomalies with higher accuracy than static, rule-based systems \cite{16}. However, traditional IDS do not generalize well to new attack variants, and modern machine learning models are highly susceptible to adversarial evasion attempts \cite{15}. Moreover, attackers can alter some features of the malicious traffic to keep the original malicious intent while changing the label to be interpreted as benign traffic by classification algorithms \cite{18}. This constitutes a major design flaw because systems that are created to detect anomalies have become targets for anomalies, which is alarming, as future use would seem unlikely.

These challenges motivate the design of detection frameworks that are immune to adversarial attacks \cite{17}. In this context, this study aims to prove that the integration of an adversarial approach as an augmentation strategy of datasets can enhance the robustness of detection systems. In this study, we use Wasserstein Generative Adversarial Networks with Gradient Penalty (WGAN-GP) to generate realistic adversarial DDoS traffic. Unlike standard GANs, WGAN-GP generates high-quality synthetic samples possessing improved training stability. WGAN-GP allows fine-tuning of adversarial traffic that replicates real attack traffic while remaining diverse. These flows are then appended to the training of machine learning models to harden them against evasion.


The primary contributions of this work include:
\begin{itemize}
    \item Comprehensive Vulnerability \& Hardening Benchmark: We quantify the extreme fragility of clean-trained RF, DNE, and Transformer architectures under feature-space evasion (Adv-5, Adv-9) and evaluate their post-augmentation recovery. 
    
    \item Protocol-Aware Feature Realizability Analysis: We provide a rigorous qualitative and structural discussion on feature dependency constraints, examining how flow-level perturbations map to actual protocol behaviors. 
    
    \item Multi-Phase Live Deployment Simulation: We evaluate model performance across a three-phase operational cycle (Before/Normal, During/Attack Peak, After/Recovery) to highlight practical failure modes like alert fatigue and conservative under-detection.
    
    \item System Resource \& Operational Profiling: We provide concrete latency, throughput, and memory consumption figures, establishing the practical feasibility of deploying Transformer-based detection at the network edge. 
    
\end{itemize}

This paper is structured as follows: Section~\ref{sec:related} reviews the literature on DDoS detection, adversarial learning, and generative approaches. The methodology, in which we present the datasets and the detection models used, is presented in Section~\ref{sec:Methodology}. The experimental setup and the results covering both baseline and enhanced models are discussed in Section~\ref{sec:Experimental}. Finally, the summary of findings, practical implications, and future research directions are presented in Section~\ref{sec:Conclusion}.

\section{Literature Review}
\label{sec:related}
DDoS detection research has evolved from static rule-based mechanisms toward deep learning models and adversarial defense frameworks. Classical machine learning classifiers---such as Decision Trees, Naïve Bayes, Support Vector Machines (SVM), and Random Forests---achieve high detection accuracy ($>99\%$) on benchmark datasets like CAIDA 2007 and CICDDoS2019 under clean conditions \cite{1}. However, these classifiers experience severe performance degradation when exposed to subtle feature perturbations.

Nevertheless, their effectiveness was limited to clean traffic and failed under adaptive adversarial conditions. To address this gap, Ali Mustapha et al. \cite{2} introduced an adversarial neural network framework that uses Generative Adversarial Networks (GANs) to create synthetic, malicious traffic flows to train and harden the detection models. Using a Long Short-Term Memory (LSTM) network as a baseline, this framework achieved up to 100\% accuracy and maintained an F1-score of around 0.99 on real traffic. Even when subjected to adversarial evasion attacks, the system maintained high detection rates (between 91.7\% and 100\%).

Chen et al. \cite{3} developed a hybrid architecture that combines a Deep Belief Network (DBN) and a Long Short-Term Memory (LSTM) network. This architecture was specifically designed for Software-Defined Networking (SDN) environments. It achieved high performance metrics: 96.55\% accuracy, 98.53\% recall, and a 97.47\% F1-score when tested on the CICDDoS2019 dataset. This hybrid framework proved robust also against perturbation attacks, maintaining over 91\% detection rates under Fast Gradient Sign Method (FGSM) adversarial testing. Shieh et al.  \cite{4} developed the Symmetric Defense GAN (SDGAN), combining adversarial attack generation with defensive classification to achieve high recall (0.977 on CIC-IDS2018) while maintaining an AUC above 0.70 under attack. Similarly, Melo et al.  \cite{5} proposed ``Anomaly-Flow,'' a multi-domain federated GAN framework achieving ROC-AUC values between 0.74 and 0.96 across heterogeneous IoT datasets. Additional advancements in robust network intrusion detection, ensemble learning, and edge-assisted threat mitigation have further demonstrated the necessity of continuous model adaptation under dynamic attack vectors \cite{kulkarni2025robust,  algaradi2024generative, chilukuri2023adversarial}.

However, it increased network communication overhead. So, existing literature demonstrates that adversarial training, particularly GAN-based data augmentation, can effectively enhance the resilience of detection models against adaptive cyberattacks. Nevertheless, model testing is mainly restricted to offline and static environments, while real-time testing is rarely addressed. Moreover, there is a lack of research addressing the practical deployment and operational integration of these models into large-scale, ISP-level network defenses.

Current DDoS mitigation is deployed across multiple infrastructure layers to ensure comprehensive coverage. At the network edge, Internet Service Providers (ISPs) and transit providers implement upstream filtering and BGP blackholing to discard malicious traffic. High-volume attacks are typically redirected to dedicated scrubbing centers for deep inspection. Cloud-based platforms offer either "always-on" or "on-demand" scrubbing, providing the elastic scalability needed for massive volumetric floods.

For localized control, enterprises utilize on-premise hardware appliances (e.g., Radware DefensePro, Fortinet FortiDDoS) to achieve real-time detection with minimal latency. Many organizations now adopt hybrid models, combining local appliances with cloud scrubbing for peak-load scalability. Furthermore, defense is integrated at the application level via Web Application Firewalls (WAFs) and behavioral analysis. Ultimately, effective mitigation relies on collaborative strategies between Autonomous System (AS) operators and victim organizations to coordinate filtering and threat intelligence.

\begin{table*}[t]
\centering
\caption{Comparative Analysis of DDoS Protection Architectures and Deployment Models}
\label{tab:comparative_analysis}
\begin{tabular}{|l|l|l|l|}
\hline
\textbf{Solution} & \textbf{Deployment} & \textbf{Strength} & \textbf{Limitation} \\ \hline
Edge / ISP Filtering & ISP edge & Early scalable filtering & Weak against Layer-7 attacks \\ \hline
Scrubbing Centers & Cloud / ISP & Handles volumetric attacks & Adds latency \\ \hline
Cloud Protection & Cloud & Highly scalable & Third-party dependency \\ \hline
On-Premise Appliances & Enterprise network & Low latency control & Limited scalability \\ \hline
Hybrid Models & Local + Cloud & Scalability + visibility & Complex management \\ \hline
Application-Level Defense & Servers / WAF & Detects Layer-7 attacks & Resource intensive \\ \hline
AI/ML Detection & IDS / Monitoring systems & Detects unknown attacks & Vulnerable to adversarial traffic \\ \hline
Proposed Framework & IDS / ISP / Cloud & Robust against evasive attacks & Retraining overhead \\ \hline
\end{tabular}
\end{table*}

Despite these developments, a critical comparative synthesis reveals three persistent limitations in state-of-the-art (SOTA) generative defense frameworks:

\begin{enumerate}

\item Synthetic Generation vs. Feature Realizability: Most GAN-based augmentation models (e.g., GADoT \cite{8}, SDGAN \cite{4}, Mustapha et al. \cite{2}) modify numerical feature vectors directly in mathematical latent space without validating whether the resulting vectors satisfy networking logic (e.g., packet rate vs. byte count consistency).

\item Static vs. Multi-Phase Deployment Testing: Prior works overwhelmingly evaluate trained models on static, offline test partitions \cite{2,4,6}. They fail to observe model behavior under operational shift across sequential traffic phases (e.g., normal baseline traffic transition into peak attack and post-attack cleanup phases).

\item Omission of Operational Overhead Metrics: While complex deep learning ensembles and GAN pipelines report high detection scores, they routinely omit inference latency, memory footprint, and throughput figures—metrics critical for high-speed edge hardware or ISP scrubbing center deployment \cite{3,5,13}. 
\end{enumerate}


To overcome these challenges, this study adopts an adversarial training framework that has the following characteristics: (i) it uses WGAN-GP to synthesize high-diversity evasive flows; (ii) it evaluates baseline ML/DL and ensemble models under both adversarial and non-adversarial regimes; and (iii) it investigates generalization and feasibility in realistic network conditions. By combining robust data augmentation with real-world testing, we move beyond simple offline accuracy to build a defense system that is truly resilient in live environments.

\section{Methodology}
\label{sec:Methodology}

\begin{figure*}[htbp]
    \centering
    \includegraphics[width=\linewidth]{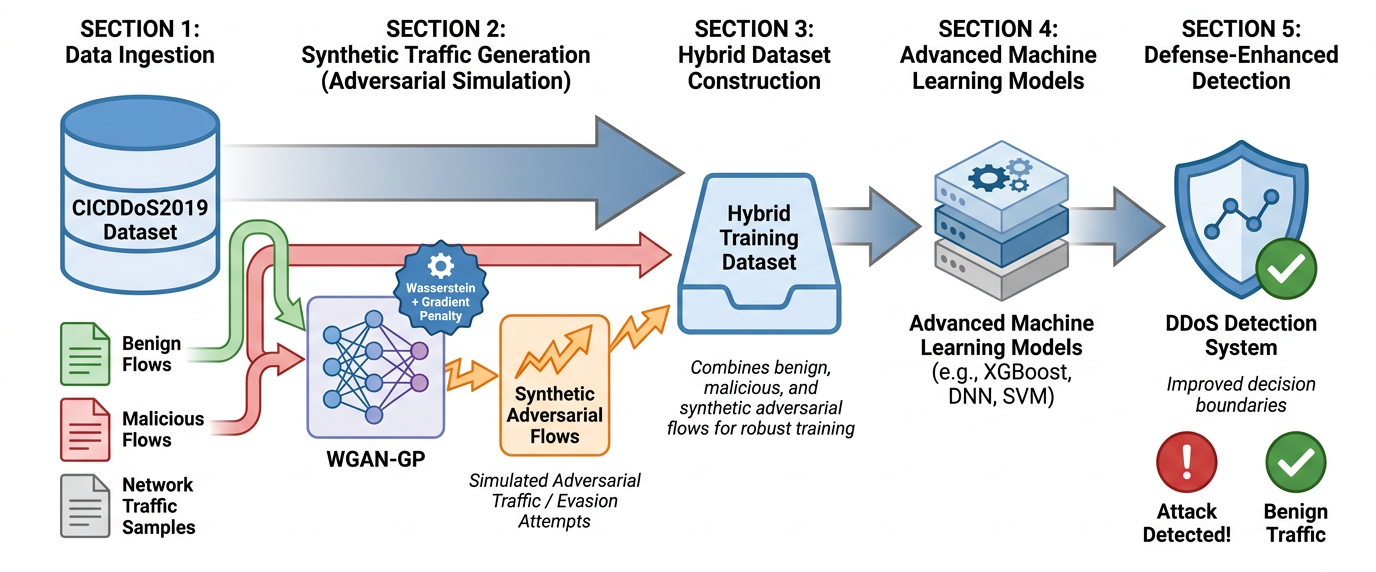}
    \caption{DDoS Attack Detection Framework Integrating Generative Adversarial Modeling with Advanced Machine Learning}
    \label{fig:ddos_framework}
\end{figure*}

This section presents the proposed framework as illustrated in Fig.~\ref{fig:ddos_framework}. The CICDDoS2019 dataset is supplemented with synthetic adversarial flows generated by WGAN-GP. These flows are combined with benign and malicious traffic to create robust hybrid datasets for training three models: The Random Forest (RF), Deep Ensemble (DNE), and Transformer (TF). The models were evaluated under both normal and adversarial conditions, including a simulated real-time deployment.


\subsection{DDoS attacks model}

We model a DDoS attack against a server as a \textit{multi-stage queuing network}, where traffic originates from a botnet and traverses a series of routers within the ISP infrastructure before reaching the target server. The network is represented as a directed graph $\mathcal{G} = (\mathcal{N}, \mathcal{L})$, where $\mathcal{N}$ denotes the set of nodes and $\mathcal{L}$ the set of links with finite bandwidth capacities $C_l$ (in Mbps) and propagation delays $d_l$ (in ms). Each router $n \in \mathcal{N}$ (last ISP) is modeled as a \textit{finite-buffer queue} with a First-In-First-Out (FIFO) discipline, characterized by:

\begin{itemize}
    \item Arrival process: legitimate traffic follows a Poisson process with rate $\lambda_{\text{legit},n}$, while attack traffic is modeled as a \textit{Markov Modulated Poisson Process (MMPP)} to capture burstiness, with state transition matrix $\mathbf{Q}$ and arrival rates $\boldsymbol{\Lambda} = [\Lambda_1, \Lambda_2, \dots, \Lambda_m]$ for $m$ attack intensity states.
    \item Service process: the service time at router $n$ is exponentially distributed with rate $\mu_n = \frac{C_n}{L_{\text{pkt}}}$, where $L_{\text{pkt}}$ is the average packet size (in bits). The service rate is capped by the link capacity $C_n$.
    \item Queue dynamics: each router has a buffer of size $K_n$ packets. The queue evolution is governed by the balance equations for an \textit{M/M/1/K} system, with steady-state probabilities $\pi_{n,j}$ for $j \in [0, K_n]$ packets in the queue. The probability of packet drops due to buffer overflow is given by:
    \begin{equation}
        P_{\text{drop},n} = \pi_{n,K_n} = \frac{\left(\frac{\lambda_{\text{total},n}}{\mu_n}\right)^{K_n} \left(1 - \frac{\lambda_{\text{total},n}}{\mu_n}\right)}{1 - \left(\frac{\lambda_{\text{total},n}}{\mu_n}\right)^{K_n + 1}},
    \end{equation}
    where $\lambda_{\text{total},n} = \lambda_{\text{legit},n} + \lambda_{\text{attack},n}$ is the aggregate arrival rate at router $n$.
    \item Transmission errors: packets are subject to transmission errors with probability $p_e$, modeled as a Bernoulli process independent of the queue state.
\end{itemize}

The last ISP router (denoted as $n_{\text{last}}$) is augmented with an \textit{AI-based DDoS detector} to classify incoming traffic as either legitimate or malicious. The detector operates as follows:
\begin{itemize}
    \item Feature extraction: for each incoming packet flow, a feature vector $\mathbf{x}_i = [f_1, f_2, \dots, f_d]$ is extracted, where $f_j$ represents features such as:
    \begin{itemize}
        \item Packet inter-arrival time.
        \item Packet size and entropy.
        \item Source/destination IP reputation scores.
        \item Flow duration and connection rate.
    \end{itemize}
    \item Classification model: a supervised machine learning model (e.g., Random Forest (RF), Deep Ensemble Neural Network (DNE), and Transformer (TF)) is trained offline on labeled datasets of normal and attack traffic. The model outputs a probability $p_{\text{malicious}} \in [0, 1]$ that a flow is part of a DDoS attack.
    \item Mitigation action: if $p_{\text{malicious}} > \theta$ (where $\theta \in [0, 1]$ is a tunable threshold), the flow is classified as malicious and discarded. Otherwise, it is forwarded to the next hop. The detector introduces an additional processing delay $D_{\text{detect}}$ (in ms) for classification.
\end{itemize}
The detection model is periodically retrained to adapt to evolving attack patterns, with a false positive rate $FPR$ and false negative rate $FNR$.

\subsection{CICDDoS2019 Dataset}
The primary dataset used in this work is CICDDoS2019 \cite{11}. It contains over 80 flow-based features extracted via CICFlowMeter, and it represents a mix of benign traffic along with several DDoS attacks (UDP Flood, HTTP Flood, SYN Flood, DNS Flood, MSSQL, LDAP, NTP, NetBIOS, SSDP, UDP-Lag, and WebDDoS attacks). The dataset is imbalanced with more attack instances than benign ones. First, we preprocessed the data by removing non-informative columns, imputing missing values, and standardizing all features. Second, we performed feature selection using the ANOVA F-test (SelectKBest), and we retained the top 20 most discriminative features for all experiments.

\subsection{Adversarial Datasets}
To generate adversarial attack traffic, we use a Wasserstein Generative Adversarial Network with Gradient Penalty (WGAN-GP). The WGAN-GP framework is composed of two neural networks engaged in a minimax game: a Generator $G$ and a Critic $C$.

The standard WGAN objective uses the Wasserstein distance (Earth-Mover distance) to measure the difference between real and generated distributions:

\begin{equation}
\min_G \max_{C \in \mathcal{L}} \mathbb{E}_{x \sim \mathbb{P}_r}[C(x)] - \mathbb{E}_{\tilde{x} \sim \mathbb{P}_g}[C(\tilde{x})]
\end{equation}

\begin{equation}
x_{adv}^{(i)} =
\begin{cases}
x_{benign}^{(j)} & \text{if } i \in \mathcal{F}_{mod} \\
\tilde{x}^{(i)} & \text{otherwise}
\end{cases}
\end{equation}

where $\mathbb{P}_r$ is the real data distribution (attack flows from CICDDoS2019), $\mathbb{P}_g$ is the generator’s distribution, and $\mathcal{L}$ is the set of 1-Lipschitz functions. To enforce the Lipschitz constraint, WGAN-GP introduces a gradient penalty term:

\begin{equation}
\lambda \mathbb{E}_{\hat{x} \sim \mathbb{P}_{\hat{x}}}[(\| \nabla_{\hat{x}} C(\hat{x}) \|_{2} - 1)^{2}]
\end{equation}

where $\hat{x}$ is a random interpolation between real and generated samples: $\hat{x} = \epsilon x + (1 - \epsilon) \tilde{x}$ with $\epsilon \sim U[0,1]$, and $\lambda$ is the penalty coefficient (set to 10 in our experiments).

The complete loss functions are:

\begin{equation}
\mathcal{L}_C = \mathbb{E}_{\tilde{x} \sim \mathbb{P}_g}[C(\tilde{x})] - \mathbb{E}_{x \sim \mathbb{P}_r}[C(x)] + \lambda \mathbb{E}_{\hat{x} \sim \mathbb{P}_{\hat{x}}}[(\| \nabla_{\hat{x}} C(\hat{x}) \|_{2} - 1)^{2}]
\end{equation}

\begin{equation}
\mathcal{L}_G = -\mathbb{E}_{\tilde{x} \sim \mathbb{P}_g}[C(\tilde{x})]
\end{equation}

\subsubsection{Architecture Specifications}

\begin{itemize}
    \item Generator $G$: Input: 100-dim noise $z \sim \mathcal{N}(0,1)$. Architecture: Dense(128) $\rightarrow$ ReLU $\rightarrow$ Dense(256) $\rightarrow$ ReLU $\rightarrow$ Dense(20) $\rightarrow$ Tanh. Output: Synthetic attack feature vector $\tilde{x}$.

    \item Critic $C$: Architecture: Dense(256) $\rightarrow$ LeakyReLU(0.2) $\rightarrow$ Dense(128) $\rightarrow$ LeakyReLU(0.2) $\rightarrow$ Dense(1) $\rightarrow$ Linear. Output: Scalar "realness" score.
\end{itemize}

The model was trained for 100 epochs with batch size 256 using the Adam optimizer ($\beta_1 = 0.5$, $\beta_2 = 0.999$, learning rate $= 2 \times 10^{-4}$), with the critic updated 4 times per generator update to ensure proper convergence.

\subsubsection{Adversarial Dataset Generation}
The trained generator produces synthetic attack flows $\tilde{x} \sim \mathbb{P}_g$. To create \textit{adversarial examples} that mimic benign traffic while retaining malicious functionality, we apply a targeted feature perturbation:

\begin{equation}
x_{adv}^{(i)} =
\begin{cases}
x_{benign}^{(j)} & \text{if } i \in \mathcal{F}_{mod} \\
\tilde{x}^{(i)} & \text{otherwise}
\end{cases}
\end{equation}

where $\mathcal{F}_{mod}$ is the set of modified features, and $x_{benign}^{(j)}$ is a value sampled from the empirical distribution of benign traffic for feature $i$.

We create two categories of datasets:
\begin{itemize}
    \item Hybrid Training Datasets: For adversarial augmentation, the original CICDDoS2019 training set is combined with adversarial samples where $|\mathcal{F}_{mod}| = \{2,4,6\}$ randomly selected features are modified.

    \item Adversarial Test Sets: For robustness evaluation, we generate test sets where $|\mathcal{F}_{mod}| = \{5,9\}$ of the \textit{most discriminative features} (ranked by ANOVA F-score) are perturbed, creating Adv-5 and Adv-9 datasets.
\end{itemize}

\subsection{Detection Models}
\subsubsection{Random Forest (RF)}
We implemented a Random Forest classifier as a baseline. It was configured with 100 decision trees with no explicit depth limitation. To overcome dataset imbalance, we also used balanced class weighting. 

\subsubsection{Deep Ensemble Neural Network (DNE)}
We built an ensemble of five independently trained feed-forward neural networks to optimize deep learning and keep the model stable. Each individual network has an identical architecture: Input $\rightarrow$ Dense(64, ReLU) $\rightarrow$ Dropout(0.3) $\rightarrow$ Dense(32, ReLU) $\rightarrow$ Dropout(0.3) $\rightarrow$ Dense(1, Sigmoid). We obtained final predictions by averaging the outputs of all five models.

\subsubsection{Transformer-based Model (TF)}
To capture global dependencies among traffic features, we designed a model based on the Transformer. To process the traffic data, our architecture embeds the 20-dimensional input into a 32-dimensional space. It then utilizes four attention heads to identify key feature patterns, incorporating layer normalization and residual links for better flow, and concludes with a standard feed-forward block. The output is flattened and passed through a final classification head (Dense(64, ReLU) $\rightarrow$ Dropout(0.3) $\rightarrow$ Sigmoid). We trained neural models (DNE, TF) using the Adam optimizer with binary cross-entropy loss for 20 epochs (batch size=256), with a validation size of 10\% of the training data. The proposed framework functions as an intelligent DDoS detection component designed for integration within diverse security environments, including enterprise gateways, cloud-based IDS, and ISP edge monitoring systems. By using a standardized feature-extraction pipeline, the system can process real-time flows captured directly from network infrastructure, such as routers and switches. The core of the architecture—a Transformer-based model hardened by WGAN-GP adversarial augmentation—is specifically engineered to maintain balanced performance and robustness against stealthy, evasive traffic patterns. This versatility allows the framework to operate either as a standalone module or as a resilient layer within a broader defense stack (alongside firewalls and scrubbing services), supporting both localized enterprise security and large-scale ISP operations.

\section{Experimental Setup and Results}
\label{sec:Experimental}

\subsection{Phase 1: Baseline Model Performance}
In the first phase, we focused on establishing a baseline for later comparison. We chose three models: Random Forest (RF), Deep Ensemble Neural Network (DNE), and Transformer (TF), and trained them on the clean CICDDoS2019 dataset. These models were evaluated not only using the clean dataset, but also against two adversarial sets: Adv-5 and Adv-9; these represent traffic where the top-5 and top-9 most discriminative features were modified, respectively. Since the main priority was to avoid missing any DDoS attacks, we chose Recall as the main metric. We also used the F1-score to balance Precision. The results of the evaluation on the clean dataset show excellent performance. All three models achieved Recall scores above 0.99 as shown in Table \ref{tab:baseline_clean}. These results confirm the models’ ability to identify standard DDoS patterns.

\begin{table}[h]
\centering
\caption{Baseline model performance on the clean CICDDoS2019 test set.}
\label{tab:baseline_clean}
\begin{tabular}{|c|c|c|}
\hline
\textbf{Model} & \textbf{Recall} & \textbf{F1-Score} \\ \hline
Random Forest (RF) & 0.9991 & 0.9995 \\ \hline
Deep Ensemble (DNE) & 0.9960 & 0.9980 \\ \hline
\end{tabular}
\end{table}

However, the models’ performance dropped significantly when evaluated on adversarially modified traffic (Table \ref{tab:baseline_adversarial}). This reveals a generalization gap as training models only on a clean dataset makes them unable to adapt to modified features. The RF and DNE models were the most affected by this degradation; Recall on Adv-9 dropped to 0.1036 and 0.0465, respectively. The TF model achieved a recall score of 0.2045 and demonstrated greater relative resilience compared to the other architectures.

\begin{table}[h]
\centering
\caption{Baseline model vulnerability to adversarial test sets.}
\label{tab:baseline_adversarial}
\begin{tabular}{|c|c|c|c|c|}
\hline
\textbf{Model} & \textbf{Recall (Adv-5)} & \textbf{F1 (Adv-5)} & \textbf{Recall (Adv-9)} & \textbf{F1 (Adv-9)} \\ \hline
RF & 0.5451 & 0.7056 & 0.1036 & 0.1877 \\ \hline
DNE & 0.2910 & 0.4509 & 0.0465 & 0.0888 \\ \hline
TF & 0.6504 & 0.7882 & 0.2045 & 0.3396 \\ \hline
\end{tabular}
\end{table}

This significant decline in performance illustrates that traditional models are not sufficient for modern security as they cannot adapt to evolving attacks. Consequently, they need to be strengthened through adversarial training.

\subsection{Phase 2: Robustness via Adversarial Augmentation}
To fix the weaknesses identified in the first phase, we retrained the three models using hybrid datasets which combine the original clean dataset (CICDDoS2019) with adversarial samples generated by the WGAN-GP, where 5 randomly selected features were modified. The retrained models were then re-evaluated on the same test sets (Adv-5 and Adv-9) to quantify the detection improvement. The results confirm that adversarial augmentation of the training dataset yielded a significant improvement in model performance, as shown in Table \ref{tab:augmented_performance}. The RF model achieved high Recall scores on both adversarial test sets,s representing a substantial enhancement over its Phase 1 results. The TF model showed remarkable performance, while maintaining a high Recall score of 0.8002 on the hardest evaluation (Adv-9). The DNE showed an improvement on Adv-5 but failed to sustain it on Adv-9, remaining the weakest model among the three.

\begin{table}[h]
\centering
\caption{Model performance on adversarial test sets after adversarial augmentation training.}
\label{tab:augmented_performance}
\begin{tabular}{|c|c|c|c|c|}
\hline
\textbf{Model} & \textbf{Recall (Adv-5)} & \textbf{F1 (Adv-5)} & \textbf{Recall (Adv-9)} & \textbf{F1 (Adv-9)} \\ \hline
RF & \textbf{1.0000} & \textbf{1.0000} & 0.9689 & 0.9842 \\ \hline
DNE & 0.5848 & 0.7380 & 0.1631 & 0.2805 \\ \hline
TF & 0.9994 & 0.9997 & \textbf{0.8002} & \textbf{0.8890} \\ \hline
\end{tabular}
\end{table}

The results demonstrate that WGAN-GP-based adversarial augmentation effectively strengthens DDoS detectors against feature-space evasion. The Transformer's consistent high performance across both phases highlights its inherent capacity to learn robust, generalized feature representations.

\subsection{Real-Time Traffic Evaluation}
To assess the viability of the models in a real-world scenario, we tested them in a simulation that mimics a real-time pipeline using three unlabeled traffic segments that represent realistic network conditions: The first one represents normal network operation and was used as a baseline for false positives. The second segment captures the peak of a DDoS attack, and the third one is the recovery phase with residual malicious traffic.

To mimic a realistic scenario, we performed inference flow-by-flow using only existing preprocessing tools (scaler, feature mask) without any additional training or adjustment during the simulation. The models exhibited distinct performance patterns, which reveal the necessity of balance between sensitivity and specificity.

\begin{table}[h]
\centering
\caption{Model performance on unlabeled real-time traffic segments.}
\label{tab:realtime_performance}
\begin{tabular}{|c|c|c|c|c|}
\hline
\textbf{Segment} & \textbf{Metric} & \textbf{RF} & \textbf{DNE} & \textbf{TF} \\ \hline
Before (Normal) & Accuracy & 1.00 & 0.00 & 1.00 \\ \hline
During (Attack) & Recall & 0.01 & 1.00 & 0.69 \\ \hline
After (Recovery) & Recall & 0.10 & 1.00 & 0.80 \\ \hline
\end{tabular}
\end{table}

The RF model achieved an accuracy score of 1.00 during normal operations, indicating it did not produce false alarms; however, it almost entirely failed to detect attacks during the attack peak (recall score during the attack phase was 0.01). This conservative approach makes it useless for real-world applications as a primary detector. The DNE labeled all traffic as malicious across the three periods (before, during, and after). This means the false positive rate is 100\% during normal operation, rendering it impractical due to alert fatigue. The TF maintained perfect accuracy during normal operation, resulting in zero false alarms; furthermore, it was able to detect the majority of attacks (recall score during the attack was 0.69). Moreover, it was able to detect malicious traffic during the recovery phase. This balance makes it the most suitable model for real-world deployment.

\subsection{Discussion}
By critically evaluating the results of this study, we can draw the following conclusions:

On one hand, the results of phase 1 (Table \ref{tab:baseline_clean}) demonstrate that relying exclusively on clean datasets can be misleading. While the three models achieved near-perfect accuracy, their performance decreased significantly when exposed to adversarial attacks. On the other hand, the recovery of the RF and TF models demonstrates that adversarial augmentation using WGAN-GP is an efficient strategy to enhance model generalization and resilience. By analyzing each architecture’s performance, we observe that design directly influences operational behavior:

\begin{itemize}
    \item The RF can be described as an extreme conservative; it failed to flag attacks in real-time tests.
    \item The DNE was hypersensitive; it triggered excessive false alarms.
    \item The Transformer model achieved significant improvements from adversarial training and is the most robust model for real-world deployment. Moreover, it provides a balance between sensitivity (recall) and precision, avoiding overreactions to normal traffic variations.
\end{itemize}

\section{Conclusion and Future Work}
\label{sec:Conclusion}

In this study, we evaluated the effectiveness of machine learning and deep learning models in detecting DDoS attacks in both controlled and real-world settings. The main requirement for a practical DDoS defense system is to achieve a balance between recall (to avoid missing attacks) and precision (to avoid triggering false alarms). Failing to achieve this balance makes the system ineffective for real-world deployment. The results show that Random Forest (RF) performs well on normal traffic but fails against active attacks. The Deep Ensemble (DE) models performed well at detecting attacks (high recall), but triggering too many false positives undermined their practical utility. These results demonstrate that the high performance of models on standard datasets degrades in real-world scenarios due to adversarial attacks; consequently, these models lack the adaptability to detect dynamic and changing attacks. The Transformer-based model achieved high recall and sensitivity, but it requires more tuning to achieve better precision. To move this model from laboratory environments to the real-world, we must harden it against evasion attacks by employing adversarial augmentation techniques using WGAN-GP. This strategy was successful, as evidenced by the evaluation metrics.

Future work will expand this framework along two primary vectors: Problem-Space Adversarial Testing: Implementing packet-level evasion tools (e.g., ptf-agent, craft-based packet injection) to evaluate closed-loop, protocol-constrained attack realizability in physical testbeds. Cross-Dataset Generalization: Validating the WGAN-GP pipeline across additional benchmark suites (e.g., Bot-IoT, ToN-IoT) and exploring online learning techniques for real-time model updating under dynamic zero-day attack patterns.


\bibliographystyle{IEEEtran}
\bibliography{references}

\end{document}